# Non-Contact Breath Rate Classification Using SVM Model and mmWave Radar Sensor Data

Mohammad Wassaf Ali[b], Ayushi Gupta[b], Mujeev Khan[a], and Mohd Wajid[a] (Senior Member, IEEE)
[a]Department of Electronics Engineering, Z.H.C.E.T. Aligarh Muslim University, India
[b]Department of Computer Engineering, Z.H.C.E.T. Aligarh Muslim University, India

*Abstract*—This work presents the use of frequency-modulated continuous wave (FMCW) radar technology combined with a machine learning (ML) model to differentiate between *Normal* and *Abnormal* Breath Rates (BR). The recommended system non-contactly collects data (or signals) using FMCW radar, which depends on BR. Various Support Vector Machine (SVM) kernels are used to classify the observed data into *Normal* and *Abnormal* states. Prolonged experiments show good accuracy in BR classification, confirming the model's efficacy. The best accuracy is 95% with the smallest number of support vectors in the case of the quadratic polynomial kernel.

## I. INTRODUCTION

Vital signs are essential markers of a person's health and are used extensively in the field of medicine to determine a persons' well being. The metrics, – blood pressure, breathe rate, body temperature, and heart rate, provide necessary information about the significant processes of the body. Vital signs are also important because they can help identify health problems at an early stage, which helps in proper treatment. They also play a critical role in clinical decision-making, chronic condition monitoring, and treatment effectiveness assessment [1] [2]. By using these signs to establish a baseline health status, doctors can identify substantial variances that might point to underlying health issues, which forms the foundation for better practices [1] [2]. Essentially, routine vital sign measurement is a fundamental component of patient assessment, providing thorough and ongoing health monitoring and enabling timely medical intervention whenever required. A healthy breadth rate is vital to human health because it improves metabolic processes, relaxes the body, and helps supply oxygen to cells[3]. Breathing exercises that show effectiveness can enhance general health and cardiovascular performance. In order to maintain the blood's normal oxygen saturation range of 94% to 98% (SpO2) at rest, effective breathing requires ventilation and gaseous exchange, with a respiratory rate that ranges from 12 to 20 bpm. [4].

Our study utilizes cutting-edge radar technology to indicate breath rate non-contactly. Using FMCW radar sensors, this technique picks up on minute movements of the abdomen and chest as depicted in Fig. 1 that correlate to respiration cycles. Without the need for physical touch, the device precisely calculates the BR of an individual by examining the frequency changes brought on by these motions. In a variety of therapeutic situations, this method guarantees patient comfort, lowers the danger of infection, and offers accurate respiratory monitoring. In this work we have employed a ML model for classification of BR as either *Normal* or *Abnormal* which performs with an accuracy of more than 95%.

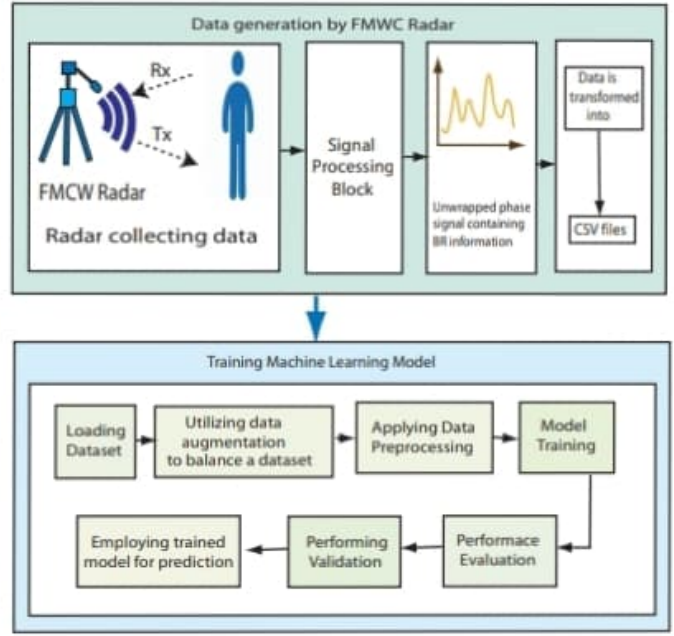

Figure 1: Overview of the proposed approach for detecting normal and abnormal breathing rates using fMCW radar and machine learning.

BR measurement has been carried out using a variety of conventional methods, such as manual counting. Physical sensors have also been used, such as chest straps and breathing belts [5]. These sensors track how the chest and belly expand and contract when breathing. Although these techniques can yield precise measures, they involve direct contact with the patient and can be obtrusive [6]. Patients may find this uncomfortable, especially if they are being monitored for long periods of time. It may also result in problems such as skin irritation or restricted movement. Furthermore, manual counting is prone to human error and might not be practical for clinical settings where continuous monitoring is needed. To measure respiratory activity without making physical contact with the patient, non contact methods for measuring breath rate have attracted a lot of attention. Prominent examples include techniques like thermal imaging [7], which measures temperature variations linked to breathing, and radar based approaches, which gauge movements of the chest wall [7]. Compared to conventional methods, these non contact techniques have a number of benefits. First of all, they do not need physical touch, lowering the possibility of cross contamination a crucial factor in situations involving infectious diseases. Secondly, they improve patient comfort since they do not require straps or adhesive sensor. Additionally, non-contact methods facilitate continuous monitoring , allowing for real time detection of respiratory abnormalities [8] [9]. This capability is crucial for early intervention and timely medical response in critical care settings. Advancements in signal processing and ML algorithms have significantly improved the accuracy and reliability of non contact BR classification, making them viable alternatives to traditional methods [10] .

Numerous studies have demonstrated the efficacy of FMCW radar in monitoring vital signs, particularly breath rate, due to its ability to detect subtle chest movements associated with breathing, enabling non-contact monitoring. For instance, J. Benny et al. developed an FMCW radar system for BR monitoring, tested in controlled environments, and highlighted its high accuracy and potential for continuous respiratory health monitoring [11]. Their study showcased a 77 GHz mmWave radar system capable of monitoring multiple subjects simultaneously without physical contact. Similarly, Abdul-Atty et al. designed a C-Band FMCW radar system specifically for BR estimation, demonstrating its accuracy in various experimental conditions [12]. This research outlined the radar's configuration and signal processing techniques tailored for breath rate monitoring. These investigations collectively demonstrate that FMCW radar is a viable non-contact technique for detecting vital signs, making it a promising technology for continuous health monitoring in both clinical and home settings.

ML techniques have revolutionized health classification by enabling the analysis of complex medical data to extract meaningful patterns and insights [13] [10] [14] [15] [16]. These models are adept at handling large datasets with numerous variables, making them invaluable for tasks such as disease diagnosis, risk prediction, and treatment outcome analysis. Various ML algorithms, including decision trees, neural networks, ensemble methods (e.g., random forests, gradient boosting), and clustering techniques, have been extensively applied in healthcare settings. These algorithms allow for the identification of intricate relationships and patterns in medical data that traditional statistical methods may overlook. By using sensor data analysis to differentiate between *Normal* BR and *Abormal* BR, machine learning helps with breathing rate health classification. Enabling real-time monitoring and predictive health insights, it processes and classifies breathing data using techniques similar to SVM.

SVM is a powerful supervised learning algorithm widely used for classification tasks in healthcare [17][18]. It works by finding an optimal hyperplane as seen in Fig. 2 that maximizes the margin between different classes in the data space. SVM's ability to handle high-dimensional data and nonlinear relationships makes it particularly suitable for analyzing medical datasets characterized by complex interactions among variables. Moreover, SVM can effectively manage small sample sizes and is robust against overfitting.

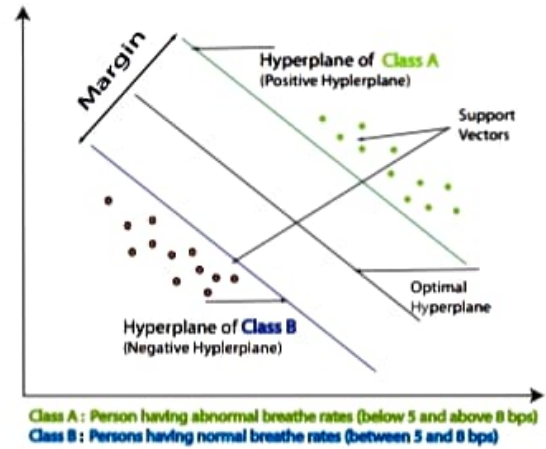

Figure 2: SVM hyplerplane maximizing class margins

Variants of SVM, including kernel SVM and Support Vector Regression (SVR), have been created to tackle distinct obstacles in the domain of health-related classification and prediction [19]. By converting the input characteristics into a higher-dimensional space using kernel functions like polynomial or radial basis function (RBF) kernels , kernel SVM improves the algorithm's handling of nonlinear data. SVM may successfully classify data with this transformation that is not linearly separable in its original form. Similar to how SVM is extended to regression tasks by SVR, accurate continuous health metrics prediction is made possible. These improvements increase SVM's adaptability and efficiency for intricate medical datasets.

To characterise breath rates as either normal or abnormal, we use SVM in our work. A decision boundary, or hyperplane, must be produced using SVM in order to partition these two groups. We accomplish this by transforming the data into a higher-dimensional space where it is easier to discern between *Normal* and *Abnormal* BR. To do this, we employ a degree two polynomial kernel. By taking this technique, the SVM architecture is able to categorise various breathing patterns with greater accuracy.

The paper is organized as follows. Section II details the study design, data gathering strategies, and analytical methods. Section III explains the experimental setup, including participant placement, radar configuration, and data processing procedures. Section IV presents the results, comparing different models based on accuracy, precision, recall, and F1 score. Finally the section V concludes the work and gives insights about the future research direction.

## II. METHODOLOGY

FMCW radar operates by transmitting a continuous signal with a frequency that varies linearly over time [20]. This signal, known as the frequency sweep or chirp, given in Eq. (1), is emitted towards the target (e.g., the subject's chest for BR classification).

$$s(t) = \cos\left[2\pi\left(f_0 t + \frac{K}{2}t^2\right)\right], \quad (1)$$

where $f_0$ is the starting frequency and $K$ is the chirp rate. The radar then receives the reflected signal, given by Eq. (2), which is delayed and Doppler-shifted due to the distance and chest wall oscillation (e.g., breathing) of the subject.

$$r(t) = \cos\left[2\pi\left(f_0\left(t - \frac{2\tau}{c}\right) + \frac{K}{2}\left(t - \frac{2\tau}{c}\right)^2\right)\right] \quad (2)$$

where $\tau$ is the round-trip delay and $c$ is the speed of light. The key principle of FMCW radar lies in measuring the phase shift between the transmitted and received signals. Figure 3 (a) demonstrates the fundamental operation of an FMCW radar, where a chirp signal is transmitted and its reflection from a target is received. A chirp, shown in Fig. 3(b), is a sinusoidal wave with a frequency that increases linearly over time. The Intermediate Frequency (IF) signal, which typically has a frequency of less than a few MHz, is created by mixing the transmitted chirp signal ($A\cos(w_{TX} \cdot t)$) with the received chirp signal ($A\cos(w_{RX} \cdot t)$), resulting in $A\cos(|(w_{RX} - w_{TX})| \cdot t)$. Where $\Delta t$ represents the chirp duration.

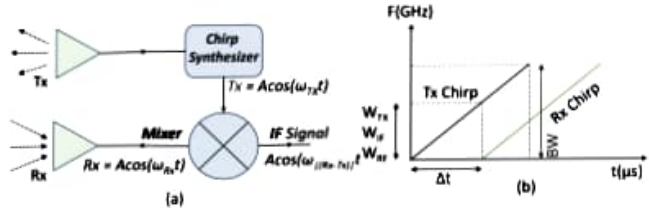

Figure 3: Basic principle of FMCW Radar Operation and chirp signal

By analyzing the frequency difference between the transmitted and received signals, known as the beat frequency, FMCW radar can accurately detect the range of the target [21]. However, the small oscillatory motion of the chest wall due to respiration can not be estimated using the beat signal frequency. To get the signal (or data) sensitive to small chest wall motion, a sequence of chirps gets transmitted, and the phase corresponding to each beat signal contains the cues of BR. This phase signal gets processed to get the unwrapped phase signal, followed by filtering. This processed signal (or data) is used in the classification model. A comprehensive dataset with diverse breath rates has been recorded from subjects.

Our approach uses SVM classifier with different kernel functions. The SVM is an algorithm for supervised learning that looks for the best hyperplane with the optimal margin between data points from different classes [22], [23]. The kernel approach transforms input data into a higher-dimensional space where it can be linearly separable, even when it is not separable in the original space. We are using three different kernal functions for comparison, i.e., Linear, Radial Basis Function (RBF), and Quadratic polynomial.

The linear kernel function calculates the similarity between two input vectors $x_i, x_j$ by computing their dot product.

$$K(\mathbf{x}_i, \mathbf{x}_j) = \mathbf{x}_i \cdot \mathbf{x}_j \quad (3)$$

The decision function determines the class of an input vector by combining the weighted sum of its features with a bias term:

$$f(X) = w \cdot x + b \quad (4)$$

classification in linear kernal is done by

$$class(x) = sgn(f(x)) \quad (5)$$

where, $w$ is the weight $b$ is the bias term.
The kernal, RBF is given by

$$K(\mathbf{x}_i, \mathbf{x}_j) = \exp\left(-\gamma \|\mathbf{x}_i - \mathbf{x}_j\|^2\right) \quad (6)$$

The following equation represents the decision function of kernal RBF

$$f(\mathbf{x}) = \sum_{i=1}^{n} \alpha_i y_i \exp\left(-\gamma \|\mathbf{x}_i - \mathbf{x}\|^2\right) + b, \quad (7)$$

classification is done by using the equation:

$$class(x) = sgn(f(x)) \quad (8)$$

where, $\mathbf{x}_i, \mathbf{x}_j$ are the input feature vectors, $\|.\|^2$ denotes the squared Euclidean distance between the vectors $\mathbf{x}_i$ and $\mathbf{x}_j$, $y_i$ represents the label or class of the i-th training sample and $\gamma$ is a hyperparameter that defines how much influence a single training example has. S The quadratic kernel function is a polynomial of degree 2 given by

$$K(x_1, z_1) = (x_1 \cdot z_1 + c)^2. \quad (9)$$

Expanding this, we get:

$$K(x_1, x_2; z_1, z_2) = (x_1 x_2 + x_2 z_2 + c)^2. \quad (10)$$

For general $n$-dimensional vectors $\mathbf{x}$ and $\mathbf{z}$, kernel function is given by

$$K(\mathbf{x}, \mathbf{z}) = \left(\sum_{i=1}^{n} x_i z_i + c\right)^2 \quad (11)$$

Expanding further,

$$K(\mathbf{x}, \mathbf{z}) = \left(\sum_{i=1}^{n} x_i z_i\right)^2 + 2c\left(\sum_{i=1}^{n} x_i z_i\right) + c^2 \quad (12)$$

The decision function for an SVM with a polynomial kernel of degree 2 is given by

$$f(x) = sgn\left(\sum_{i=1}^{m} \alpha_i y_i K(\mathbf{x}_i, \mathbf{x}) + b\right) \quad (13)$$

where $K(\mathbf{x_i}, \mathbf{x})$ is the polynomial kernel function of degree 2, given by

$$K(\mathbf{x_i}, \mathbf{x}) = (\mathbf{x_i} \cdot \mathbf{x} + c)^2 \qquad (14)$$

Substituting the kernel function into the decision function, we get:

$$f(x) = \text{sgn}\left(\sum_{i=1}^{m} \alpha_i y_i (\mathbf{x_i} \cdot \mathbf{x} + c)^2 + b\right) \qquad (15)$$

where, $\alpha_i$ are the Lagrange multipliers, $y_i$ are the class labels, $\mathbf{x}_i$ are the support vectors (28 in this work), $\mathbf{x}$ is the input vector, $c$ is the bias term, and $b$ is the intercept term.

We explored linear, quadratic and RBF kernels to determine the optimal model for our dataset [24] [25]. The dataset was divided into training and validation sets to evaluate model performance. We selected the most effective model for breath rate classification by comparing different SVM kernels' performance. The next stage is to train and validate the machine learning models after the dataset has been processed and unwrapped signals (or data) have been extracted. The dataset was split into validation and training sets. The training set is used to optimize the model's parameters, whereas the validation set evaluates the model's accuracy and capacity for generalization. We tested with various SVM kernels, such as linear, quadratic polynomial, and RBF; their performance is as depicted in Table I.

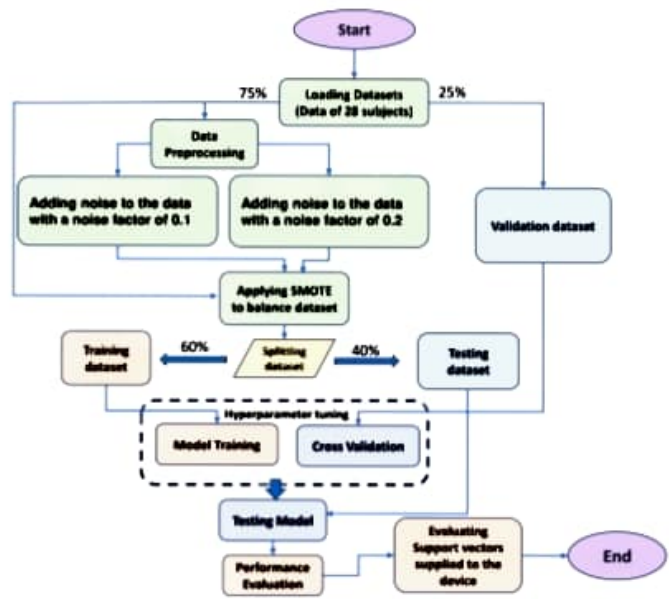

Figure 4: Diagramatic representation of model training

### III. EXPERIMENTAL SETUP

In this study, we analyzed breathing rate patterns in a diverse group of individuals using an FMCW radar system, illustrated in Fig. 1. To ensure reliable and consistent signal reflection, each participant was positioned precisely 1 meter from the FMCW radar setup. The subjects were exposed to a sequence of chirp signals emitted by the radar, which interacted with their bodies to produce reflections. These reflections were recorded and transformed into unwrapped phase signals, each timed to exact intervals of 25.6 seconds per subject. The normal range of BR is 5 to 8 breaths per 25.6 seconds, with abnormal rates falling outside this range. The experimental environment was strictly controlled to prevent signal contamination from external sources, ensuring no other individuals were present in front of the radar during data gathering. For data collection, we employed the TI-AWR1642BOOST radar module, operating in the 77-81 GHz frequency range. Our setup operates with a start frequency of 77 GHz and a stop frequency of 79.982 GHz, providing an RF bandwidth of 2998.2 MHz. The chirp duration is set to 100 $\mu s$, with a chirp period of 50 ms, and the chirp slope is 29.982 MHz/$\mu s$. The system collects 512 chirps and 512 ADC samples of IF signal per chirp. These IF signals corresponding to each chirp are processed to obtain a phase signal. Due to phase wrapping ambiguity, phase unwrapping is performed on the extracted phase. The resulting unwrapped phase signal is stored in CSV file, which is then used as data for further model training. ML models were built on top of this dataset, represented in Fig. 4.

To increase the volume of training samples, data augmentation was applied by adding noise with noise factors of 0.1 and 0.2 [26]. Additionally, the Synthetic Minority Oversampling Technique (SMOTE) was used to balance the dataset between *Normal* and *Abnormal* classes [27]. The augmented data was fed into an SVM with a different kernel function for ML and testing. The performance of the trained model was verified using unseen data.
below 5 and above 8 breaths in 25.6 seconds

### IV. RESULTS

The classification results demonstrate that the proposed system can effectively distinguish between *Normal* and *Abnormal* BR. The SVM classifier, utilizing a quadratic polynomial kernel, achieved high accuracy with a lesser number of support vectors in identifying abnormal BR. Class *Abnormal BR* represents individuals with unhealthy BR ( i.e. below 5 and above 8 breaths in 25.6 seconds) lying above the hyperplane, while Class *Normal BR* represents individuals with healthy BR (i.e. 5 to 8 breaths in 25.6 seconds) lying below the hyperplane.

The classifier's performance has been evaluated using the matrices, namely, accuracy, precision, recall, and F1 score as defined in Eqs. (16), (17), (18) and (19), respectively.

$$\text{Accuracy} = \frac{TP + TN}{TP + TN + FP + FN} \qquad (16)$$

where $TP$ is true positive, $TN$ is true negative, $FP$ is false positive, and $FN$ is false negative. Additionally, precision, recall, and F1-scores provide insights into the classifier's performance.

$$\text{Precision} = \frac{TP}{TP + FP} \qquad (17)$$

$$\text{Recall} = \frac{TP}{TP+FN} \quad (18)$$

$$\text{F1-score} = 2 \cdot \frac{\text{Precision} \cdot \text{Recall}}{\text{Precision} + \text{Recall}} \quad (19)$$

Table I: Performance comparison of Models based on accuracy, precision, recall, and F1 score.

| Classifier | Accuracy | Precision | Recall | F1 Score | Support Vectors |
|---|---|---|---|---|---|
| Linear | 86% | 95% | 77% | 87% | 29 |
| RBF | 95% | 98.89% | 92.3% | 96% | 61 |
| Quadratic | 95% | 94% | 94% | 94% | 28 |

Table I gives the results for SVM classifiers with three different kernel functions: Linear, RBF, and Quadratic. The SVM model with linear kernel attains an accuracy of 86%, whereas the Quadratic and RBF models achieve an accuracy of 95%. Although the RBF model performs well with relatively higher precision (98%) and an F1 score (96%) compared to the Quadratic model, number of support vectors is a key factor in model selection. The RBF model uses 61 support vectors, while the Quadratic model uses only 28. Fewer support vectors indicate low computational complexity of the model.

Figure 5: Confusion Matrix of the Final Model

The final model selected for the classification of *Normal* and *Abnormal* BR is a SVM with a quadratic kernel. Fig 5 presents the performance of the final model. Table I represents the accuracy (95%), precision (94%), recall (94%) and F1 Score (94%) of Quadratic kernal SVM. Fig 5 represents a confusion matrix that compares the target values actually attained with the model's predictions. These results collectively demonstrate the SVM classifier's capability in accurately categorizing respiratory health conditions, ensuring reliable classification outcomes for real-time health monitoring applications.

## V. CONCLUSION

This paper presents a non-contact solution for classifying *Normal* and *Abnormal* BR using ML approaches. In this work, we employed an FMCW radar to collect vital information and integrated that data with a ML model. We utilized an SVM classifier with multiple kernels, i.e. linear, RBF, and quadratic. Evaluating them based on performance and computational efficiency. Among all the evaluated kernels and models, the SVM with a quadratic kernel yielded the best results in terms of low computation, without compromising accuracy and other performance metrics. Future work could involve deploying this model on a hardware platform or edge device to demonstrate portability and real-time application in clinical settings.

ACKNOWLEDGEMENT

The authors would like to acknowledge Chips-to-Startup (C2S) program, Ministry of Electronics and Information Technology (MeitY), Govt. of India.

REFERENCES

[1] Anthony D Harries, Rony Zachariah, Anil Kapur, Andreas Jahn, and Donald A Enarson. The vital signs of chronic disease management. *Transactions of the Royal Society of Tropical Medicine and Hygiene*, 103(6):537–540, 2009.

[2] A. Raji, P. Golda Jeyasheeli, and T. Jenitha. Iot based classification of vital signs data for chronic disease monitoring. In *2016 10th International Conference on Intelligent Systems and Control (ISCO)*, pages 1–5, 2016.

[3] Michelle A Cretikos, Rinaldo Bellomo, Ken Hillman, Jack Chen, Simon Finfer, and Arthas Flabouris. Respiratory rate: the neglected vital sign. *Medical Journal of Australia*, 188(11):657–659, 2008.

[4] Sarah H Annesley Barry Hill. Monitoring respiratory rate in adults, 2021.

[5] Duarte Dias and João Paulo Silva Cunha. Wearable health devicesâvital sign monitoring, systems and technologies. *Sensors*, 18(8):2414, 2018.

[6] Carlo Massaroni, Andrea Nicolò, Daniela Lo Presti, Massimo Sacchetti, Sergio Silvestri, and Emiliano Schena. Contact-based methods for measuring respiratory rate. *Sensors*, 19(4):908, 2019.

[7] J Michael Lloyd. *Thermal imaging systems*. Springer Science & Business Media, 2013.

[8] Michael H Li, Azadeh Yadollahi, and Babak Taati. A non-contact vision-based system for respiratory rate estimation. In *2014 36th Annual International Conference of the IEEE Engineering in Medicine and Biology Society*, pages 2119–2122. IEEE, 2014.

[9] Carlo Massaroni, Daniela Lo Presti, Domenico Formica, Sergio Silvestri, and Emiliano Schena. Non-contact monitoring of breathing pattern and respiratory rate via rgb signal measurement. *Sensors*, 19(12):2758, 2019.

[10] Qi An, Saifur Rahman, Jingwen Zhou, and James Jin Kang. A comprehensive review on machine learning in healthcare industry: classification, restrictions, opportunities and challenges. *Sensors*, 23(9):4178, 2023.

[11] Jewel Benny, Pranjal Mahajan, Srayan Sankar Chatterjee, Mohd Wajid, and Abhishek Srivastava. Design

and measurements of mmwave fmcw radar based non-contact multi-patient heart rate and breath rate monitoring system. In *2023 IEEE Biomedical Circuits and Systems Conference (BioCAS)*, pages 1–5, 2023.

[12] Mohammad Mohammad Abdul-Atty, Ahmed Sayed Ismail Amar, and Mohamed Mabrouk. C-band fmcw radar design and implementation for breathing rate estimation. *Advances in Science, Technology and Engineering Systems Journal*, 5(5):1299–1307, 2020.

[13] Jian Ping Li, Amin Ul Haq, Salah Ud Din, Jalaluddin Khan, Asif Khan, and Abdus Saboor. Heart disease identification method using machine learning classification in e-healthcare. *IEEE access*, 8:107562–107582, 2020.

[14] Umair Muneer Butt, Sukumar Letchmunan, Mubashir Ali, Fadratul Hafinaz Hassan, Anees Baqir, and Hafiz Husnain Raza Sherazi. Machine learning based diabetes classification and prediction for healthcare applications. *Journal of healthcare engineering*, 2021(1):9930985, 2021.

[15] Yash Pratap Singh, Aham Gupta, Devansh Chaudhary, Mohd Wajid, Abhishek Srivastava, and Pranjal Mahajan. Hardware deployable edge ai solution for posture classification using mmwave radar and low computation machine learning model. *IEEE Sensors Journal*, 2024.

[16] Pranjal Mahajan, Devansh Chaudhary, Mujeev Khan, Mohammed Hammad Khan, Mohd Wajid, and Abhishek Srivastava. A point cloud-based non-intrusive approach for human posture classification by utilizing 77 ghz fmcw radar and deep learning models. In *2024 IEEE International Symposium on Circuits and Systems (ISCAS)*, pages 1–5, 2024.

[17] SVM Vishwanathan and M Narasimha Murty. Ssvm: a simple svm algorithm. In *Proceedings of the 2002 International Joint Conference on Neural Networks. IJCNN'02 (Cat. No. 02CH37290)*, volume 3, pages 2393–2398. IEEE, 2002.

[18] Anthony Fleury, Michel Vacher, and Norbert Noury. Svm-based multimodal classification of activities of daily living in health smart homes: sensors, algorithms, and first experimental results. *IEEE transactions on information technology in biomedicine*, 14(2):274–283, 2009.

[19] Mariette Awad, Rahul Khanna, Mariette Awad, and Rahul Khanna. Support vector regression. *Efficient learning machines: Theories, concepts, and applications for engineers and system designers*, pages 67–80, 2015.

[20] Andrew G Stove. Linear fmcw radar techniques. In *IEE Proceedings F (Radar and Signal Processing)*, volume 139, pages 343–350. IET, 1992.

[21] Mujeev Khan, Pranjal Mahajan, Gani Nawaz Khan, Devansh Chaudhary, Jewel Benny, Mohd Wajid, and Abhishek Srivastava. Design and implementation of fpga based system for object detection and range estimation used in adas applications utilizing fmcw radar. In *2024 IEEE International Symposium on Circuits and Systems (ISCAS)*, pages 1–5, 2024.

[22] Sheng Liu and Na Jiang. Svm parameters optimization algorithm and its application. In *2008 IEEE International Conference on Mechatronics and Automation*, pages 509–513. IEEE, 2008.

[23] Pádraig Cunningham, Matthieu Cord, and Sarah Jane Delany. Supervised learning. In *Machine learning techniques for multimedia: case studies on organization and retrieval*, pages 21–49. Springer, 2008.

[24] Yujun Yang, Jianping Li, and Yimei Yang. The research of the fast svm classifier method. In *2015 12th international computer conference on wavelet active media technology and information processing (ICCWAMTIP)*, pages 121–124. IEEE, 2015.

[25] Shunjie Han, Cao Qubo, and Han Meng. Parameter selection in svm with rbf kernel function. In *World Automation Congress 2012*, pages 1–4. IEEE, 2012.

[26] Francisco J Moreno-Barea, Fiammetta Strazzera, José M Jerez, Daniel Urda, and Leonardo Franco. Forward noise adjustment scheme for data augmentation. In *2018 IEEE symposium series on computational intelligence (SSCI)*, pages 728–734. IEEE, 2018.

[27] Alberto Fernández, Salvador Garcia, Francisco Herrera, and Nitesh V Chawla. Smote for learning from imbalanced data: progress and challenges, marking the 15-year anniversary. *Journal of artificial intelligence research*, 61:863–905, 2018.